# A knowledge-based intelligent system for control of dirt recognition process in the smart washing machines

Mohsen Annabestani [1,*], Alireza Rowhanimanesh [2], Akram Rezaei [3]. Ladan Avazpour [4], Fatemeh Sheikhhasani [5]

1. Department of Electrical Engineering, Sharif University of Technology, Tehran, Iran.
2. Department of Electrical Engineering, University of Neyshabur, Neyshabur, Iran.
3. Department of Computer engineering, Azad University of Mashhad, Mashhad, Iran.
4. Division of Applied Mathematics, Brown University, Providence, RI, USA
5. Bahar Institute of higher education, Mashhad, Iran.

*Abstract*— **In this paper, we propose an intelligence approach based on fuzzy logic to modeling human intelligence in washing clothes. At first, an intelligent feedback loop is designed for perception-based sensing of dirt inspired by human color understanding. Then, when color stains leak out of some colored clothes the human probabilistic decision making is computationally modeled to detect this stain leakage and thus the problem of recognizing dirt from stain can be considered in the washing process. Finally, we discuss the fuzzy control of washing clothes and design and simulate a smart controller based on the fuzzy intelligence feedback loop.**

## I. Introduction

The process of washing clothes is a non-linear process which many factors influence the performance including the value of detergent powder, water temperature, dirt of clothes and clothes stains. This process should be controlled, to achieve a desired performance and efficiency. For example, how much detergent powder is required? What temperature is suitable? How long should the washing process be continued? The answers highly depend on the amounts of dirt and stain of clothes as well as the type of clothes. When human washes clothes, he/she intelligently controls the process by getting feedback from clothes dirt and stain which are usually understood by looking at the effluent color. Using this perception based feedback as well as prior knowledge about washing clothes process; human determines the value of powder detergent, water temperature and wash time. In other words, human accurately controls this process by perception-based feedback and perception-based knowledge rather than numerical methods and mathematical equations. In this paper, we aim to model this perception-based control system computationally. First, we apply fuzzy logic to computationally model human understanding of clothes dirt by observing the effluent color. This process usually has the most important feedback in washing clothes procedure. It should be mentioned that the proposed approach is general enough to be used for computational modeling of other perception-based feedbacks such as the type of clothes. Then, we offer a method based on fuzzy logic and linguistic probability to recognize dirt from stain in effluent when color stains out of some colored clothes. Finally, we discuss how human knowledge about control of washing clothes can be computationally modeled.

Zadeh has introduced the concept of fuzzy sets in 1965, a mathematical model of uncertainty that results from a non-sharp boundary between the objects [1]. A few years later, he discussed the fuzzy decision-making to explain the vagueness in nature which is helpful in various area, namely, the design of the smart refrigerators and intelligent washing machines, etc. [2]. Several authors have discussed fuzzy theory's applications. For example in [3] a method for control of washing machine using Neuro-fuzzy modeling has been proposed. Besides we can find a couple of other fuzzy approaches related to the washing machine in [4-7]. A nonlinear identification method has been suggested by [8-10] employing Adaptive Neuro-Fuzzy Inference System (ANFIS).

Moreover, in [11], an accurate dynamic nonlinear black-box model was proposed to foretell the mechanical displacement features of IPMC actuators. In other works, evaluating the quality of the step response with the help of fuzzy criterion has been negotiated by the authors of this paper [12-14]. We can also find other applications of fuzzy in literature, for example, Khalili and Kasaei [15] applied fuzzy logic to object modeling for Multi-camera Correspondence. Ahmed et al. [16] proposed an approach based on fuzzy logic to content-based image retrieval. Moghani [17] used fuzzy logic in color naming. Butzke et al. [18] considered color classification and logo segmentation in automatic recognition of vehicle attributes using fuzzy computing. Annabestani and Saadatmand [19] proposed a fuzzy based Image binarization method to extract text from color document images. Shamir proposed an approach to perception-based color segmentation using fuzzy logic [20]. Xiaoling and Kanglin [21] considered the application of fuzzy logic in content-based image retrieval. Beside engineering application, fuzzy theory is a very important field in the mathematical sciences, for example, fuzzy approach widely uses in fuzzy differential equations [22, 23], fuzzy inequalities [24], etc.

* Corresponding author: Mohsen Annabestani, e-mail : annabetany@gmail.com

## II. PRELIMINARY CONCEPTS

### A. Effluent Sampling

In the process of washing clothes, amount of dirt in clothes determine from effluent color. During a washing process, effluent colors change from dark to nearly bright. So, the effluent color must be sampled through the process. In this paper, we use image processing for this feedback. First, clothes effluent comes into a little container with a cube shape as shown in Figure 1. The front side of the container must be made of thin colorless glass, and the other three sides must be purely white. Also, a small oscillation must be applied to the effluent in the container to avoid deposition. Then, using a CCD image sensor, an image is taken from effluent in the container. This process is repeated periodically through the washing process. It should be mentioned that the image acquisition environment must be nearly stationary which implies that the amount of light, noise, and distortion is almost constant and pre-determined.

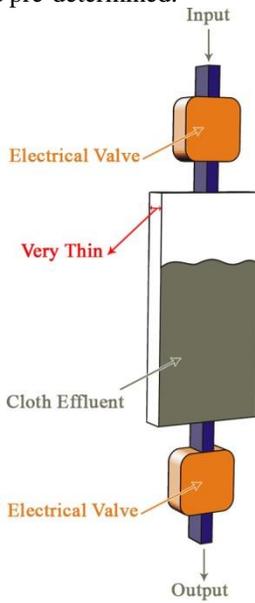

**Fig.1: Cubic Container for effluent of cloth**

### B. The Concept of Dominant Color

In this paper, what we mean by the image is the effluent image that usually has a simple texture, small color variance and thus only one dominant color. Roughly speaking, the dominant color is a color that we assign to an image in our mind when looking at it. The concept of dominant color plays an essential role in many cognitive functions of a human vision system. Since the effluent image has small color variance and the uncertainty of image acquisition environment is low, a simple mean can be used to obtain the dominant color. Since the output of CCD sensors is usually in RGB color space,

Equations (1,2,3) display how R, G and B components of dominant color can be calculated through averaging of all pixels where CCD returns an image with x and y pixels. In these equations, $\sum ij\ R$ is the sum of Red components of all pixels and similarly for Green and Blue, and x× y is the total number of pixels. The resulted dominant color is (Rd, Gd, Bd). To demonstrate the efficiency of these equations, Figure 2 shows several examples for calculating dominant color using these equations. It is crucial to note that these equations are applicable only if the given image has small color variance and the environment of image acquisition has low uncertainty. In continue of this paper, wherever we say effluent color, we mean dominant color of the effluent image.

$$Rd = \frac{\sum R_{ij}}{x \times y} \quad (1)$$

$$Gd = \frac{\sum G_{ij}}{x \times y} \quad (2)$$

$$Bd = \frac{\sum B_{ij}}{x \times y} \quad (3)$$

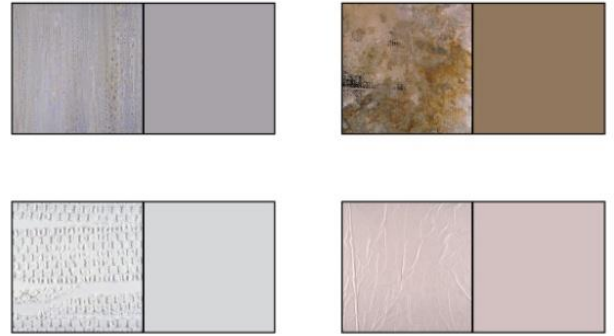

**Fig . 2: Four textures and their dominant colors.**

## III. INTELLIGENT SENSE OF DIRT

The most important feedback in washing clothes is dirt of clothes. In this section, we use the Mamdani fuzzy system to model human-like understanding of dirt based on expert knowledge computationally. For this purpose, we provide a knowledge base including linguistic if-then rules from dry cleaning experts. A summary of the significant results that we have achieved from expert's knowledge is explained below. In most of the processes of washing clothes, effluent color has a nearly constant Hue. During the washing process and over the time, effluent color is going to be clear and more transparent. Since the background of the sampling container is pure white, the color of white means that the clothes are clean enough and the washing process can be stopped. The color of white that can be ideally considered as a criterion for cleanliness of washed clothes has the maximum Intensity (100%) and the minimum Saturation (0%). Since the effluent color has a nearly constant Hue and its brightness only depends on Intensity and Saturation, the inputs of the fuzzy system select as the values of Intensity (0% - 100%) and Saturation (0% - 100%) of the dominant color of the effluent image. The output of a fuzzy system is the percent of clothes dirt (0% - 100%) which can be accurately approximated from these inputs. The fuzzy system is responsible for approximating dirt percent of clothes from Intensity and Saturation of effluent color. We represent the knowledge of dry-cleaning experts by 36 fuzzy rules. Figures 3(a,b,c), 4(a,b) and 5 indicate membership functions, fuzzy rule base and surface of this fuzzy system, respectively. Figure 4 displays how the Intensity-Saturation space is partitioned using fuzzy linguistic values defined in Figures 3 (a) and 3 (b). In Figure 3 (c), six linguistic values are assigned to the dirt of clothes where O_MAX describes the maximum of dirt percent (about 90–100 %) which equals to black effluent color and O_MIN describes the maximum of cleanliness percent (dirt percent of about 0–10 %) that equals to white effluent color. From the viewpoint of input-output mapping, Figure 5 shows the corresponding surface of the designed fuzzy system that is simulated in MATLAB Fuzzy Logic Toolbox. As expected, the surface is close enough to the linguistic knowledge base of dry-cleaning experts shown in Figure 5. For example, it shows that for minimal values of Intensity, the output has its maximum value which means that the percent of dirt is the highest. Also, the minimum percent of dirt occurs where Intensity is the maximum and Saturation is the minimum. Through several simulation examples in the next section, we demonstrate the efficiency of the designed fuzzy system.

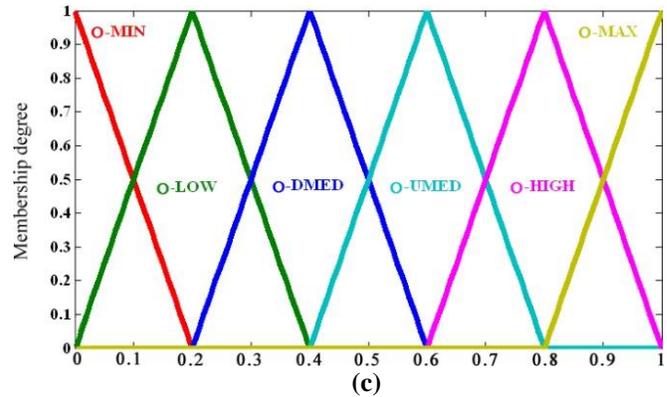

Fig . 3: (a) Saturation, (b) Intensity and (c) : Dirt Spectrum

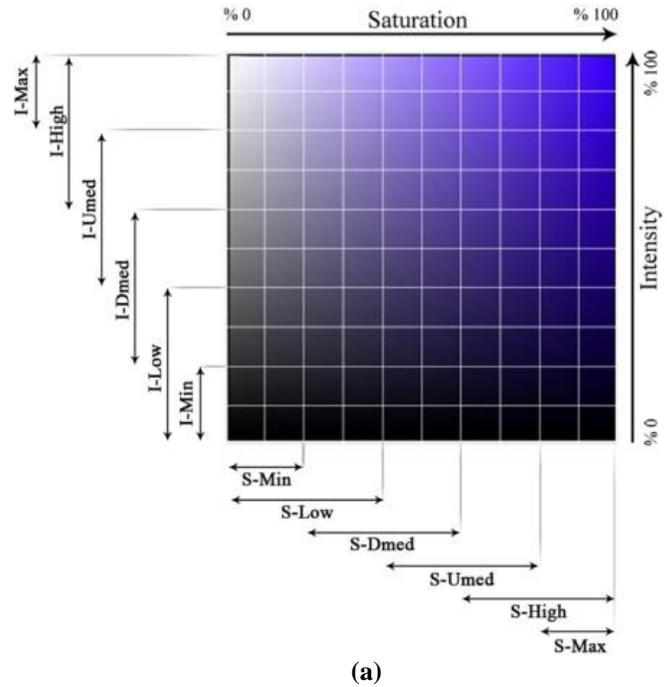

(a)

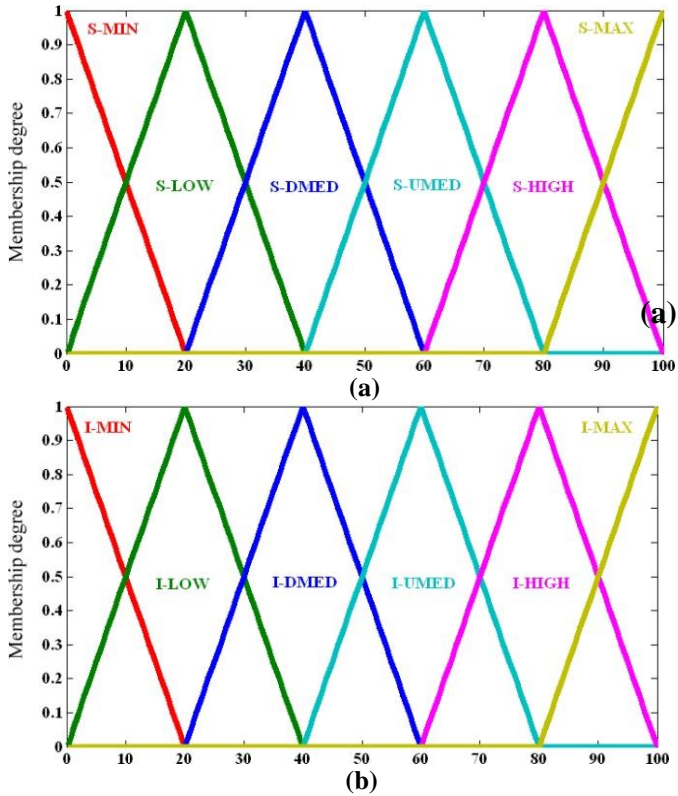

| I S | I-Min | I-Low | I-DMed | I-UMed | I-High | I-Max |
|---|---|---|---|---|---|---|
| S-Min | O-Max | O-High | O-Umed | O-Low | O-Min | O-Min |
| S-Low | O-Max | O-High | O-Umed | O-Dmed | O-Low | O-Min |
| S-DMed | O-Max | O-High | O-Umed | O-Dmed | O-Dmed | O-Low |
| S-UMed | O-Max | O-Max | O-Umed | O-Umed | O-Umed | O-Dmed |
| S-High | O-Max | O-Max | O-High | O-High | O-Umed | O-Dmed |
| S-Max | O-Max | O-Max | O-High | O-High | O-High | O-Umed |

(b)

Fig. 4: (a) Knowledge Base for a Typical Constant Hue, (b) Fuzzy Rule Base.

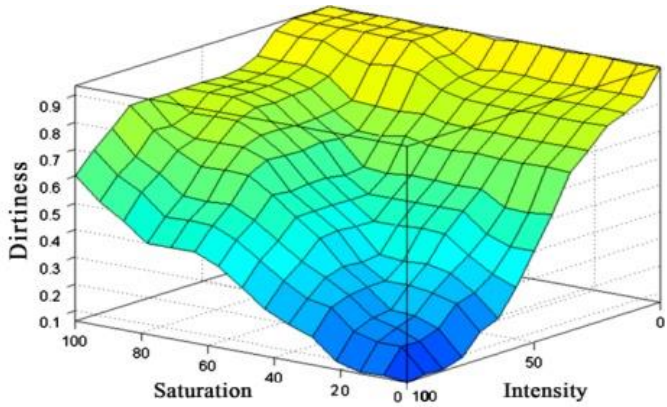

**Fig. 5: Fuzzy system surface**

## A. Simulation Examples I

Several examples as like as real effluent color are simulated in MATLAB to confirm the performance of the designed fuzzy system. Each sample sequence simulates a 900 seconds interval of a washing process where the effluent color changes during the process. Here, we suppose that the clothes have no stain. The Intensity and Saturation of each sample apply to the designed fuzzy system as an input vector, and the fuzzy system returns the dirt percent as an output. Figure.6 shows the first sample where the effluent color changes from almost black (very large dirt percent) to almost white (very small dirt percent) and the output of the designed fuzzy system can successfully approximate the expected dirt percent. Figure.7 simulates another situation when suddenly an amount of dirt is added to the process at 400s. This figure shows that the fuzzy system can adequately detect this sudden change of dirt percent. Generally, the results demonstrate that the designed fuzzy system can closely model human understanding of dirt percent.

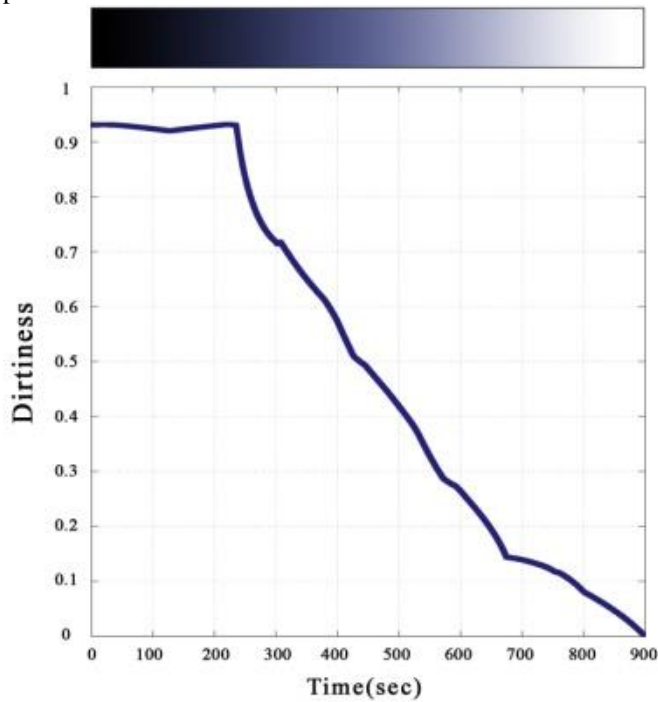

**Fig. 6: Simulation Example: Dirt Percent as Fuzzy System Output.**

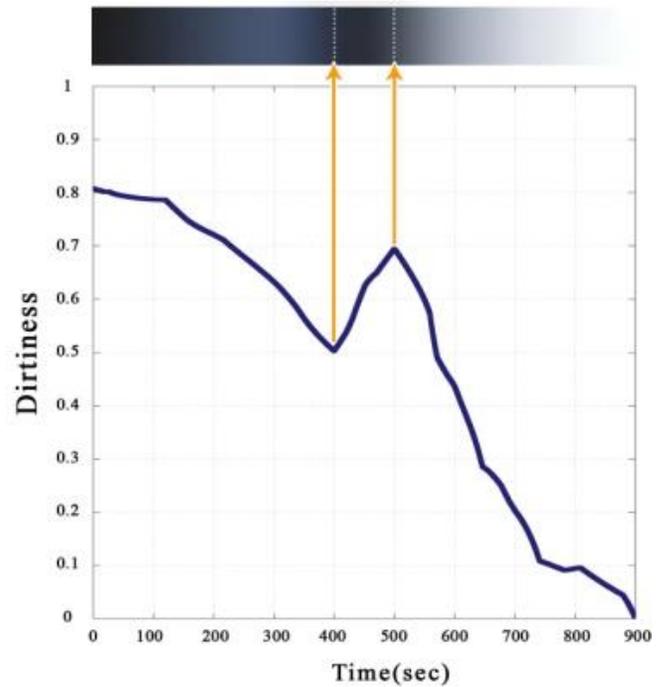

**Fig. 7: Simulation Example: Dirt Percent as Fuzzy System Output**

## IV. RECOGNITION OF DIRT FROM STAIN IN THE PRESENCE OF COLORED CLOTHES

One essential point in washing clothes is recognition of dirt from stain when color stains out of some colored clothes. This status often occurs in the presence of new colored clothes. Usually, dry-cleaning experts can recognize dirt from stain correctly. The designed fuzzy system in the previous section cannot recognize dirt from stain. In the presence of colored clothes when the color that stains out of them, the effluent color might never become white. So, the designed fuzzy system in the last section thinks that the dirt still exists, This means that the washing process must be continued while the clothes have been already cleaned. Indeed, this inaccuracy leads to waste of energy, time and cost. In this section, we aim to model human intelligence in recognizing dirt from stain computationally.

It should be noted that in most cases especially in dry-cleaning, a human does not already know if the given clothes stain or not. He must recognize it during the washing process from effluent color. Thus, his decision is probabilistic rather than deterministic. Human has prior knowledge of the dirt color in his mind. For example, most of the dry cleaning experts consider a color between gray and miry green as the color of dirt. To distinguish dirt from stain, human compares the effluent color with the expected dirt color. After several observations during the washing process, he can identify the difference with a value of probability of the truth of the effluent color related to dirt or stain. The amount of this probability highly influences the washing strategy. For example, if this probability is high, it means that it is very likely that color stains out of some clothes, thus high temperatures for water should not be used and the clothes should be quickly dried and separated after washing.

In this section, we aim to model this probabilistic decision making of the human computationally using a fuzzy system and the concept of linguistic probability. According to a knowledge base that we achieved experimentally from several dry-cleaning experts, we have considered the color spectrum displayed in Figure.8 as the possible colors for dirt of clothes. The membership function that indicates in Figures.9 shows the possibility of distribution over the possible dirt color. It should be mentioned that the proposed methods of this paper are general and not only limited to these examples. Therefore, figures like Figure.8 must be edited for real-world applications. We design a fuzzy system that takes the HSI components of dominant effluent color as an input vector, compares it with the expected dirt color shown in Figure.8, and finally returns the probability of stain as an output. Figures.9 shows the membership functions defined over the domains of three inputs. These membership functions describe the possibility distribution of Figure.8 in HSI color space. The fuzzy system output is the probability of stain. Figure.10 displays linguistic values defined over probability space [0 1]. The high output value means the high stain probability. If the dominant effluent color is in the possible region of Figure.8 with large possibility (membership) degree, then the fuzzy system output must be negligible which implies that the stain probability is low.

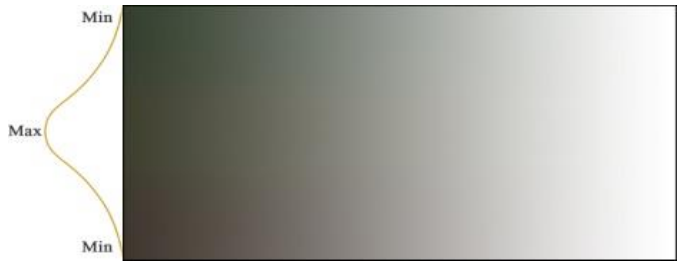

Fig. 8: The most dominant areas of hypothetical dirty in the human mind

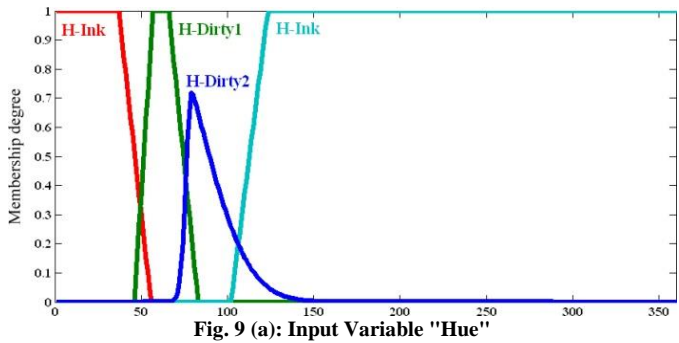

Fig. 9 (a): Input Variable "Hue"

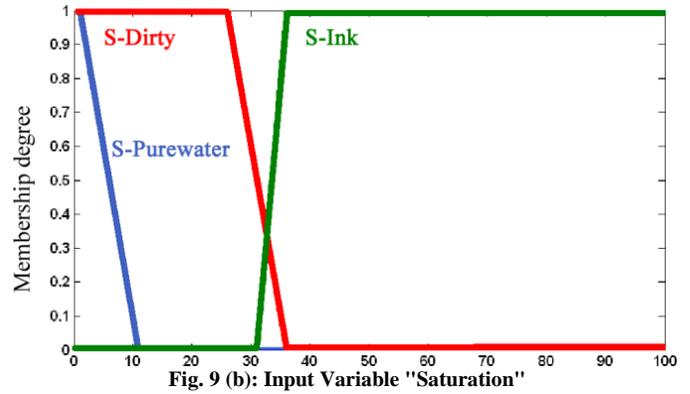

Fig. 9 (b): Input Variable "Saturation"

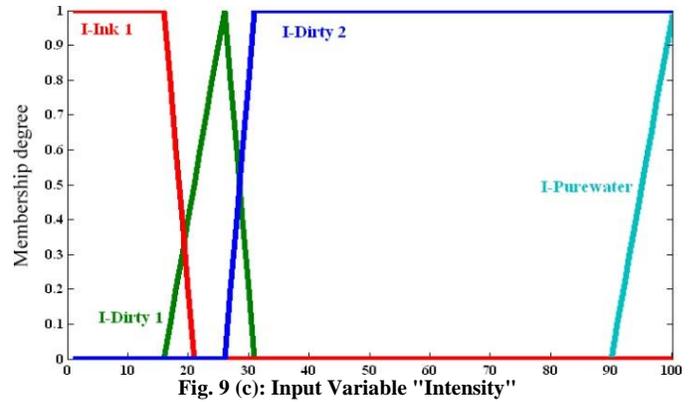

Fig. 9 (c): Input Variable "Intensity"

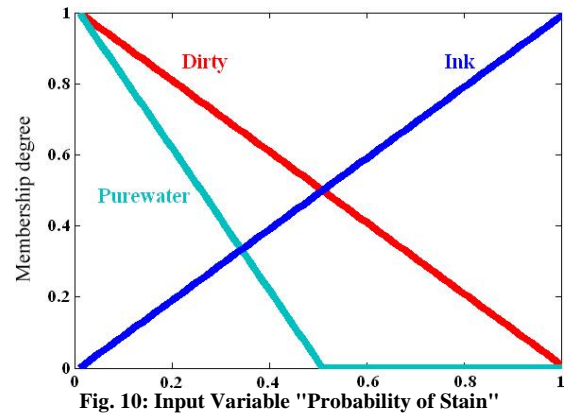

Fig. 10: Input Variable "Probability of Stain"

Here, we describe the expert knowledge of fuzzy controller only by three simple fuzzy if-then rules as follows:

1. *If Hue is H-ink and Saturation is S-ink and Intensity is I-ink then Probability of Stain is High*
2. *If Hue is H-dirty1 and Saturation is S-dirty and Intensity is I-ink1 then Probability of Stain is Low*
3. *If Hue is H-dirty2 and Saturation is S-dirty and Intensity is I-ink2 then Probability of Stain is Low*

Figures.11 show the surfaces of this fuzzy system. It is easy to find that these surfaces are compatible with the mentioned fuzzy rule base.

It should be noted that depending on the given application; different rule bases can be defined. Also, in addition to the inputs as mentioned above, other inputs such as time can be considered to estimate the probability of stain more accurately.

For example, if stain color is in the range of dirt color, then the designed fuzzy system must be modified and other factors such as time must be used to approximate the probability of stain.

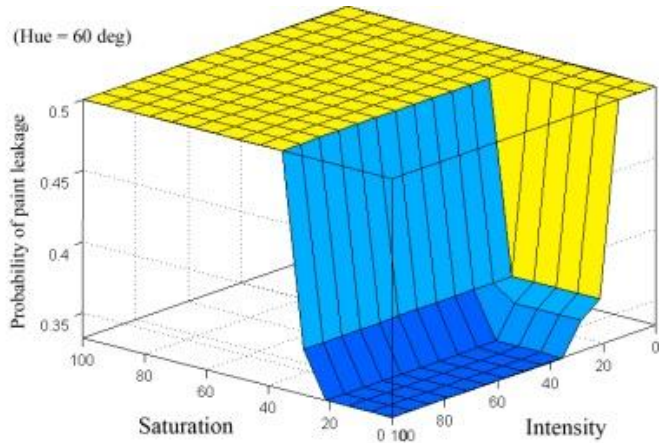

Fig. 11(a): Output Surface (Hue = 60)

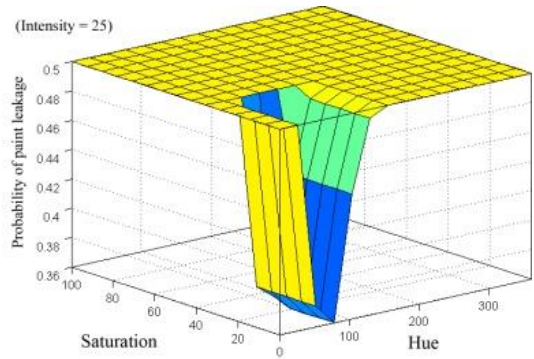

Fig. 11(b): Output Surface (Intensity = 25)

### A. SIMULATION EXAMPLE II

To demonstrate the efficiency of the designed fuzzy system, Figure.12 shows a simulation example designed in MATLAB as similar as examples of Section III.A. The simulated effluent color (input vector of the fuzzy system) thoroughly stains (not dirt) until about 800s (Figure.13). After 800s the color is located in the possible range of dirt colors shown in Figure.8. The output of the fuzzy system that is the probability of stain is shown as a plot in Figure.12. As expected, this plot demonstrates that the designed fuzzy system can successfully approximate the likelihood of stain.

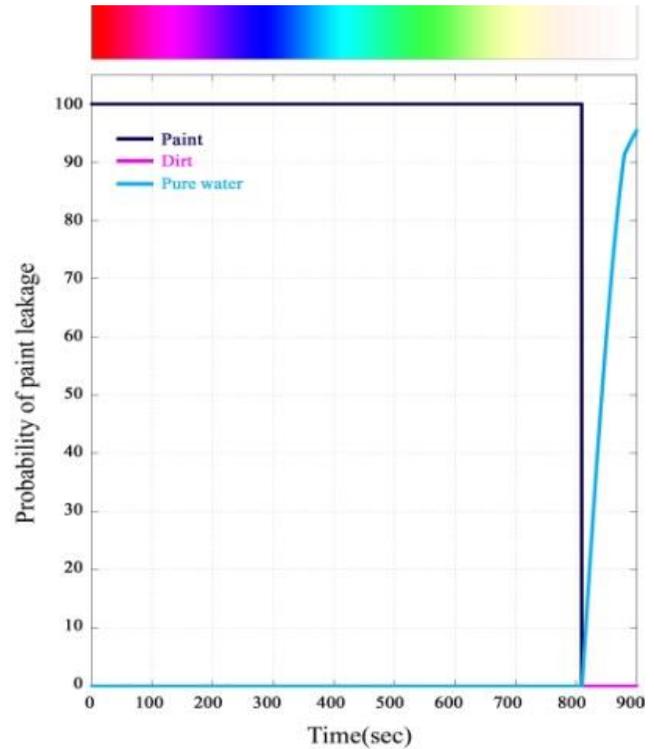

Fig.12: Simulation Example: Probability of Stain in the presence of Colored Effluent

### B. Cleanliness Criterion

In Section 3, we discuss that when the clothes are going to be clean, the effluent color is going to be clear and more evident and thus the dirt percent is going to be lower and lower. But if color stains out of some colored clothes, even after all clothes are clean, the effluent is still colored. Here, we show that the previous cleanliness criterion can be still used. Suppose color stains out of some colored clothes; This means that the probability of stain is one. At the end of washing process when the clothes are cleaned, although effluent is still colored, its Saturation and Intensity values are very low (about 5%) and very high (about 95%), respectively. The amount of Hue is not crucial since its effect can be ignored. If these values are applied to the designed fuzzy system of Section 3, it returns 0.01 that is very close to zero. Thus, we can still use the fuzzy system of Section 3 since very small values of dirt percent can be considered as cleanliness criterion.

### V. CONCLUSION

The In this paper, we propose a general approach based on fuzzy logic to computationally modeling of human intelligence in washing clothes. Both problems of intelligent feedback and intelligent control are considered. It should be noted that the mentioned methods are general enough to be used, with minor changes, for different applications in the context of dry-cleaning. Generally, intelligent control of washing clothes can efficiently reduce the waste of energy, time and cost and also it can improve the quality of washing. The proposed approach of this paper can be considered as a computational tool for intellectualization of dry-cleaning industries, especially in large scales.